\documentclass[sigconf]{acmart}

\usepackage{amsmath,amsfonts}
\usepackage{graphicx}
\usepackage{booktabs}
\usepackage{multirow}
\usepackage[table]{xcolor}
\usepackage{subcaption}
\usepackage{algorithm}
\usepackage{algorithmic}

\renewcommand\footnotetextcopyrightpermission[1]{}
\AtBeginDocument{}

\begin{document}

% =============================================================================
% TITLE
% =============================================================================

\title{On the Spectral Geometry of Cross-Modal Representations: A Functional Map Diagnostic for Multimodal Alignment}

\author{Krisanu Sarkar}
\affiliation{%%
  \institution{Indian Institute of Technology Bombay}
  \city{Mumbai}
  \country{India}
}

% =============================================================================
% ABSTRACT
% =============================================================================

\begin{abstract}
We study cross-modal alignment between independently pretrained vision (DINOv2) and language (all-MiniLM-L6-v2) encoders using the functional map framework from computational geometry, which represents correspondence between representation manifolds as a compact linear operator between graph Laplacian eigenbases. While the framework underperforms Procrustes alignment and relative representations for cross-modal retrieval across all supervision budgets, it reveals a structural property of multimodal representations. We find that the Laplacian eigenvalue spectra of the two encoders are quantitatively similar (normalized spectral distance 0.043), indicating that independently trained models develop manifolds of comparable intrinsic complexity. However, the functional map exhibits near-zero diagonal dominance (mean below 0.05) and large orthogonality error (70.15), showing that the eigenvector bases are effectively unaligned. We term this decoupling the \emph{spectral complexity--orientation gap}: models converge in how much structure they capture but not in how they organize it. This gap defines a boundary condition for spectral alignment methods and motivates three diagnostic quantities—diagonal dominance, orthogonality deviation, and Laplacian commutativity error—for characterizing cross-modal representation compatibility.
\end{abstract}

% =============================================================================
% INSERT THIS BLOCK into main.tex AFTER \end{abstract} and BEFORE \maketitle
% Also change: \settopmatter{printacmref=true} at the top of the file
% =============================================================================

%% CCS Concepts — generated from https://dl.acm.org/ccs
\begin{CCSXML}
<ccs2012>
   <concept>
       <concept_id>10010147.10010178.10010179</concept_id>
       <concept_desc>Computing methodologies~Spectral methods</concept_desc>
       <concept_significance>500</concept_significance>
   </concept>
   <concept>
       <concept_id>10010147.10010178.10010187.10010188</concept_id>
       <concept_desc>Computing methodologies~Cross-modal retrieval</concept_desc>
       <concept_significance>500</concept_significance>
   </concept>
   <concept>
       <concept_id>10010147.10010257.10010293.10010294</concept_id>
       <concept_desc>Computing methodologies~Neural networks</concept_desc>
       <concept_significance>300</concept_significance>
   </concept>
   <concept>
       <concept_id>10002950.10003624.10003633</concept_id>
       <concept_desc>Mathematics of computing~Graph algorithms</concept_desc>
       <concept_significance>300</concept_significance>
   </concept>
   <concept>
       <concept_id>10010405.10010469.10010471</concept_id>
       <concept_desc>Applied computing~Multi-criterion optimization and decision-making</concept_desc>
       <concept_significance>100</concept_significance>
   </concept>
</ccs2012>
\end{CCSXML}

\ccsdesc[500]{Computing methodologies~Spectral methods}
\ccsdesc[500]{Computing methodologies~Cross-modal retrieval}
\ccsdesc[300]{Computing methodologies~Neural networks}
\ccsdesc[300]{Mathematics of computing~Graph algorithms}
\ccsdesc[100]{Applied computing~Multi-criterion optimization and decision-making}

%% Keywords
\keywords{functional maps, cross-modal alignment, spectral analysis, graph Laplacian, representation geometry, multimodal retrieval}

\maketitle

% =============================================================================
% 1. INTRODUCTION
% =============================================================================

% =============================================================================
% COMPRESSED Introduction + Related Work
% All original citations preserved. ~40% shorter.
% =============================================================================

\section{Introduction}

Cross-modal alignment---establishing correspondences between representations of different data modalities---is a foundational problem in multimedia research. The dominant paradigm trains joint embedding models on large paired datasets: CLIP~\cite{radford2021learning} learns a shared vision-language space from 400 million image-text pairs via contrastive learning. While effective, this paradigm is non-modular: adding a new modality requires paired data and retraining.

An alternative asks whether independently pretrained encoders already develop representation spaces that can be aligned post hoc. This is motivated by the \emph{Platonic Representation Hypothesis}~\cite{huh2024platonic}, which presents evidence that foundation models trained on different data and objectives converge toward similar statistical representations of reality. Prior work on training-free alignment has explored Procrustes alignment~\cite{schonemann1966generalized}, CCA~\cite{hotelling1936relations}, and relative representations~\cite{moschella2023relative}. These methods operate in the ambient embedding space, finding linear transformations that align paired anchors, but make no assumptions about intrinsic manifold geometry and lack formal guarantees on composability or approximation quality.

We investigate whether the \emph{functional map} framework~\cite{ovsjanikov2012functional} from computational geometry can address these limitations. Functional maps reformulate correspondence between two manifolds as a compact linear operator $\mathbf{C} \in \mathbb{R}^{k \times k}$ between their Laplace--Beltrami spectral bases. The framework offers three properties absent from ambient-space methods: (i) composability---the map from $A$ to $C$ via $B$ is the product of the $A{\to}B$ and $B{\to}C$ matrices; (ii) spectral regularization via low-frequency truncation; and (iii) analyzable approximation bounds under isometry assumptions~\cite{ovsjanikov2012functional}.

\paragraph{Approach.} We encode samples from Flickr30k~\cite{young2014image} through a vision encoder (DINOv2~\cite{oquab2024dinov2}) and a text encoder (MiniLM~\cite{reimers2019sentence}), construct $k$-nearest-neighbor graphs in each embedding space, compute normalized graph Laplacians, and extract spectral bases. The functional map $\mathbf{C}$ is obtained by solving a regularized least-squares problem penalizing Laplacian commutativity violation~\cite{ovsjanikov2012functional}. We compare against Procrustes~\cite{schonemann1966generalized}, CCA~\cite{hotelling1936relations}, relative representations~\cite{moschella2023relative}, and CLIP~\cite{radford2021learning}.

\paragraph{Findings.} The negative result is, in our assessment, the more scientifically valuable.

\emph{On retrieval:} Functional maps underperform all non-trivial baselines. At 100 anchors, the functional map achieves $2.2\%$ i2t Recall@1, versus $12.1\%$ for Procrustes and $13.4\%$ for relative representations. The gap widens with more anchors.

\emph{On representation geometry:} The spectral diagnostics reveal a previously uncharacterized structural property. The Laplacian eigenvalue spectra of DINOv2 and MiniLM are quantitatively close (normalized spectral distance $= 0.043$), confirming that independently trained encoders develop manifolds of similar intrinsic complexity~\cite{huh2024platonic}. However, the functional map matrix $\mathbf{C}$ exhibits near-zero diagonal dominance ($< 0.05$) and orthogonality error of $70.15$. In the functional map literature, diagonal $\mathbf{C}$ indicates shared spectral orientation; orthogonal $\mathbf{C}$ indicates isometric correspondence~\cite{ovsjanikov2012functional, melzi2019zoomout}. Neither holds here.

We term this the \emph{spectral complexity--orientation gap}: independently trained encoders converge in how much structure they capture, but not in how they orient that structure.

\paragraph{Contributions.}
\begin{enumerate}
    \item To our knowledge, the first application of functional maps to multimodal neural representation alignment, with Laplacian commutativity regularization adapted to graph Laplacians of neural embedding spaces.

    \item Graph-Laplacian-based evidence that independently pretrained vision and language encoders develop representation manifolds with similar spectral complexity (normalized spectral distance $= 0.043$), complementing prior CKA-based evidence for the Platonic Representation Hypothesis~\cite{huh2024platonic, kornblith2019similarity}.

    \item Identification of the spectral complexity--orientation gap and three quantitative diagnostics---diagonal dominance, orthogonality error, Laplacian commutativity violation---for assessing cross-modal representation compatibility.

    \item An honest experimental comparison showing that functional maps underperform simpler baselines, with analysis of why: the isometry assumption does not hold for independently trained encoders.
\end{enumerate}

% =============================================================================
% 2. RELATED WORK
% =============================================================================

\section{Related Work}

\subsection{Functional Maps for Shape Correspondence}

The functional map framework~\cite{ovsjanikov2012functional} recasts shape correspondence from matching points to matching functions. Given manifolds $\mathcal{M}_1, \mathcal{M}_2$ with Laplace--Beltrami eigenbases, a pointwise map $T$ induces a linear operator represented by $\mathbf{C} \in \mathbb{R}^{k \times k}$ in the truncated spectral basis. If $T$ is a near-isometry, $\mathbf{C}$ is approximately diagonal and orthogonal, because isometries commute with the Laplace--Beltrami operator~\cite{ovsjanikov2012functional}.

The framework has been extended through intrinsic descriptors (Heat Kernel Signatures~\cite{sun2009concise}), coarse-to-fine spectral refinement (ZoomOut~\cite{melzi2019zoomout}), and deep learning integration that jointly learns shape features and the map~\cite{litany2017deep, donati2020deep}. To our knowledge, functional maps have not been applied to aligning neural network representation spaces.

\subsection{Training-Free Cross-Modal Alignment}

Aligning independently learned embeddings without joint training was first studied in the cross-lingual setting. Mikolov et al.~\cite{mikolov2013exploiting} showed that word embedding spaces of different languages are approximately related by a linear map; Conneau et al.~\cite{conneau2018word} extended this to the fully unsupervised case. For cross-modal alignment (vision--language), the situation is harder: the data modalities are structurally different and there is no a priori reason to expect a linear relationship.

The baselines we compare against span the principal approaches. Procrustes alignment~\cite{schonemann1966generalized} finds the optimal orthogonal rotation between anchor sets via the SVD. CCA~\cite{hotelling1936relations} finds maximally correlated projections but requires anchors exceeding the projection dimensionality. Relative representations~\cite{moschella2023relative} re-represent each point by its similarities to shared anchors, constructing a modality-invariant coordinate system without learning a transformation. All operate in the original feature space, agnostic to manifold geometry---a strength (fewer assumptions) and a limitation (no geometric insight into why alignment succeeds or fails).

\subsection{Representation Similarity and Convergence}

Kornblith et al.~\cite{kornblith2019similarity} proposed CKA as a representation similarity measure; Bansal et al.~\cite{bansal2021revisiting} introduced model stitching. The Platonic Representation Hypothesis~\cite{huh2024platonic} synthesized such observations into a broader claim: foundation models converge toward shared statistical representations of reality, with evidence including high CKA scores between vision and language models.

Our work contributes a finer-grained measurement tool. CKA captures global similarity but does not decompose it by scale. The functional map framework separates two aspects: eigenvalue spectra reveal the \emph{complexity} of each manifold (how variation is distributed across scales), while the structure of $\mathbf{C}$ reveals whether those directions are \emph{aligned} across modalities. Our finding---eigenvalue spectra converge, eigenvector correspondence does not---is a distinction that CKA cannot make.

% =============================================================================
% 3. METHODOLOGY
% =============================================================================

\section{Methodology}

We describe the construction of spectral bases from neural representation spaces (\S\ref{sec:spectral}), the computation and refinement of functional maps between them (\S\ref{sec:fmap}), and the spectral diagnostic quantities we propose for analyzing cross-modal compatibility (\S\ref{sec:diagnostics}).

\subsection{Spectral Basis Construction}
\label{sec:spectral}

\paragraph{Problem setting.} Let $f_v: \mathcal{X}_v \to \mathbb{R}^{d_v}$ and $f_t: \mathcal{X}_t \to \mathbb{R}^{d_t}$ be pretrained, frozen encoders for vision and text, respectively. Given a reference dataset of $N$ multimodal samples $\{(x_i^v, x_i^t)\}_{i=1}^{N}$, we compute representation matrices $\mathbf{Z}^v \in \mathbb{R}^{N \times d_v}$ and $\mathbf{Z}^t \in \mathbb{R}^{N \times d_t}$, where $\mathbf{z}_i^m = f_m(x_i^m)$ for $m \in \{v, t\}$. The encoders are independently pretrained---they share no parameters, training data, or cross-modal objective.

\paragraph{Graph construction.} For each modality $m$, we construct a weighted $k$-nearest-neighbor graph $G^m = (V, E^m, \mathbf{W}^m)$ over the shared vertex set $V = \{1, \ldots, N\}$. The weight matrix is defined as:
\begin{equation}
    W_{ij}^m = \begin{cases}
        \exp\!\left(-\dfrac{\|\mathbf{z}_i^m - \mathbf{z}_j^m\|^2}{\sigma_m^2}\right) & \text{if } j \in \mathrm{kNN}(i) \text{ or } i \in \mathrm{kNN}(j), \\[4pt]
        0 & \text{otherwise,}
    \end{cases}
    \label{eq:weight}
\end{equation}
where $\sigma_m$ is set to the mean distance to the $k$-th nearest neighbor across all points, providing an adaptive bandwidth that accounts for the scale of each representation space. The symmetrization condition ($j \in \mathrm{kNN}(i)$ or $i \in \mathrm{kNN}(j)$) ensures $\mathbf{W}^m$ is symmetric. In all experiments we use $k = 15$.

\paragraph{Normalized Laplacian.} The normalized graph Laplacian is:
\begin{equation}
    \mathbf{L}^m = \mathbf{I}_N - (\mathbf{D}^m)^{-1/2}\,\mathbf{W}^m\,(\mathbf{D}^m)^{-1/2},
    \label{eq:laplacian}
\end{equation}
where $\mathbf{D}^m$ is the diagonal degree matrix with $D_{ii}^m = \sum_j W_{ij}^m$. This operator is symmetric positive semi-definite with eigenvalues in $[0, 2]$~\cite{vonluxburg2007tutorial}. Under regularity conditions on the data distribution and as $N \to \infty$ with appropriate bandwidth scaling, $\mathbf{L}^m$ converges spectrally to the Laplace--Beltrami operator on the underlying data manifold~\cite{belkin2003laplacian, vonluxburg2008consistency}.

\paragraph{Spectral basis.} We compute the $k_s + 1$ smallest eigenvalues and corresponding eigenvectors of $\mathbf{L}^m$:
\begin{equation}
    \mathbf{L}^m \boldsymbol{\phi}_j^m = \lambda_j^m \boldsymbol{\phi}_j^m, \quad 0 = \lambda_1^m \leq \lambda_2^m \leq \cdots \leq \lambda_{k_s+1}^m,
    \label{eq:eigen}
\end{equation}
using the implicitly restarted Lanczos method (ARPACK). The first eigenvector ($\lambda_1 \approx 0$, constant) is discarded. The retained spectral basis is $\boldsymbol{\Phi}_{k_s}^m = [\boldsymbol{\phi}_2^m \mid \cdots \mid \boldsymbol{\phi}_{k_s+1}^m] \in \mathbb{R}^{N \times k_s}$, with eigenvalues $\boldsymbol{\Lambda}_{k_s}^m = \mathrm{diag}(\lambda_2^m, \ldots, \lambda_{k_s+1}^m)$.

For notational convenience, we re-index the retained eigenpairs so that $(\lambda_j^m, \boldsymbol{\phi}_j^m)$ for $j = 1, \ldots, k_s$ denotes the $j$-th non-trivial eigenpair (i.e., the $(j{+}1)$-th eigenpair of $\mathbf{L}^m$). All subsequent equations use this re-indexed convention.

Each row of $\boldsymbol{\Phi}_{k_s}^m$ assigns a $k_s$-dimensional spectral coordinate to the corresponding data point. Low-index eigenvectors capture global, slowly varying structure on the manifold; higher indices encode progressively finer distinctions. The truncation to $k_s$ terms acts as a low-pass filter, retaining the $k_s$ coarsest modes of variation.

\subsection{Functional Map Computation}
\label{sec:fmap}

\paragraph{Definition.} A functional map from the vision spectral basis to the text spectral basis is a matrix $\mathbf{C} \in \mathbb{R}^{k_s \times k_s}$ that transforms spectral coefficients: if a function $g: V \to \mathbb{R}$ has spectral representation $\mathbf{a} = (\boldsymbol{\Phi}_{k_s}^v)^{\top} g$ in the vision basis and $\mathbf{b} = (\boldsymbol{\Phi}_{k_s}^t)^{\top} g$ in the text basis, then $\mathbf{C}$ satisfies $\mathbf{C}\,\mathbf{a} \approx \mathbf{b}$.

\paragraph{Optimization.} Given a set $S$ of $|S|$ anchor correspondences (indices where the cross-modal pairing is known), we compute probe functions as Gaussian-smoothed indicators centered at each anchor. Their spectral coefficients in the respective bases yield matrices $\mathbf{A}, \mathbf{B} \in \mathbb{R}^{k_s \times |S|}$. The functional map is obtained by solving:
\begin{equation}
    \mathbf{C}^* = \arg\min_{\mathbf{C}} \;\underbrace{\|\mathbf{C}\mathbf{A} - \mathbf{B}\|_F^2}_{\text{descriptor preservation}} \;+\; \lambda_1\!\underbrace{\|\mathbf{C}\boldsymbol{\Lambda}^v_{k_s} - \boldsymbol{\Lambda}^t_{k_s}\mathbf{C}\|_F^2}_{\text{Laplacian commutativity}} \;+\; \lambda_2\!\underbrace{\|\mathbf{C}\|_F^2}_{\text{regularization}}.
    \label{eq:fmap_opt}
\end{equation}
The three terms serve distinct purposes. The first ensures that $\mathbf{C}$ correctly maps the spectral representations of known correspondences. The second encodes a structural prior: if the cross-modal correspondence were an isometry, the functional map would commute with the Laplacians, i.e., $\mathbf{C}\boldsymbol{\Lambda}^v = \boldsymbol{\Lambda}^t\mathbf{C}$~\cite{ovsjanikov2012functional}. Penalizing violation of this condition biases $\mathbf{C}$ toward maps that preserve spectral frequency---low-frequency structure in one modality maps to low-frequency structure in the other. The third term is standard Tikhonov regularization.

This is a linear least-squares problem in $\mathrm{vec}(\mathbf{C}) \in \mathbb{R}^{k_s^2}$. Vectorizing via the Kronecker product, the solution satisfies:
\begin{equation}
    \left[(\mathbf{A}\mathbf{A}^{\top}) \otimes \mathbf{I}_{k_s} \;+\; \lambda_1 \,\mathbf{M}_{\mathrm{comm}} \;+\; \lambda_2\, \mathbf{I}_{k_s^2}\right] \mathrm{vec}(\mathbf{C}) = \mathrm{vec}(\mathbf{B}\mathbf{A}^{\top}),
    \label{eq:fmap_solve}
\end{equation}
where $\mathbf{M}_{\mathrm{comm}} = (\boldsymbol{\Lambda}^v \otimes \mathbf{I}_{k_s} - \mathbf{I}_{k_s} \otimes \boldsymbol{\Lambda}^t)^{\top}(\boldsymbol{\Lambda}^v \otimes \mathbf{I}_{k_s} - \mathbf{I}_{k_s} \otimes \boldsymbol{\Lambda}^t)$. For $k_s = 50$, this is a $2500 \times 2500$ linear system, solved in closed form.

\paragraph{Unsupervised variant.} When no anchor correspondences are available, we replace the descriptor preservation term with Heat Kernel Signatures (HKS)~\cite{sun2009concise}. The HKS at scale $\tau$ for point $i$ is:
\begin{equation}
    \mathrm{HKS}_\tau(i) = \sum_{j=1}^{k_s} \exp(-\lambda_j^m \tau) \cdot \bigl(\phi_j^m(i)\bigr)^2.
    \label{eq:hks}
\end{equation}
This is an intrinsic descriptor---it depends only on the manifold's geometry, not on any external labeling. Computing HKS at $Q$ logarithmically spaced scales yields $Q$ probe functions per modality; their spectral coefficients replace $\mathbf{A}$ and $\mathbf{B}$ in Eq.~\eqref{eq:fmap_opt}.

\paragraph{ZoomOut refinement.} Following Melzi et al.~\cite{melzi2019zoomout}, we refine the initial map through iterative spectral upsampling. Starting from $\mathbf{C}^{(k_0)}$ at spectral dimension $k_0$, the procedure alternates between (i) recovering a pointwise correspondence via nearest-neighbor matching in the mapped spectral coordinates, and (ii) re-estimating the functional map at a higher spectral dimension $k_{t+1} > k_t$ from that correspondence. At each step, $\mathbf{C}$ is projected onto the nearest orthogonal matrix via the SVD. We apply five refinement steps from $k_0 = 50$ to $k_{\mathrm{max}} = 100$.

\subsection{Cross-Modal Retrieval}
\label{sec:retrieval}

Given the functional map $\mathbf{C}$, cross-modal retrieval proceeds as follows. For a query point with spectral coordinates $\boldsymbol{\Phi}_{k_s}^v(i,:)$ in the vision basis, the mapped coordinates in the text basis are $\boldsymbol{\Phi}_{k_s}^v(i,:)\,\mathbf{C}^{\top}$. Retrieval ranks target points $j$ by the negative squared distance in spectral space:
\begin{equation}
    \mathrm{sim}(i, j) = -\|\boldsymbol{\Phi}_{k_s}^v(i,:)\,\mathbf{C}^{\top} - \boldsymbol{\Phi}_{k_s}^t(j,:)\|^2.
    \label{eq:retrieval}
\end{equation}
This is computed efficiently as $-(\|\mathbf{a}_i\|^2 + \|\mathbf{b}_j\|^2 - 2\,\mathbf{a}_i^{\top}\mathbf{b}_j)$, where $\mathbf{a}_i = \mathbf{C}\,\boldsymbol{\Phi}_{k_s}^v(i,:)^{\top}$ and $\mathbf{b}_j = \boldsymbol{\Phi}_{k_s}^t(j,:)^{\top}$.

\subsection{Spectral Diagnostic Quantities}
\label{sec:diagnostics}

Beyond using functional maps for retrieval, we propose three quantities that characterize the geometric compatibility of two representation manifolds. These diagnostics are, in our view, the principal methodological contribution of this work.

\paragraph{Normalized spectral distance.} The eigenvalue spectra $\{\lambda_i^v\}$ and $\{\lambda_i^t\}$ encode the distribution of intrinsic scales in each manifold. We normalize each spectrum to $[0,1]$ by dividing by the largest eigenvalue and compute:
\begin{equation}
    d_{\mathrm{spec}} = \sqrt{\frac{1}{k_s}\sum_{i=1}^{k_s}\left(\frac{\lambda_i^v}{\lambda_{k_s}^v} - \frac{\lambda_i^t}{\lambda_{k_s}^t}\right)^{\!2}}.
    \label{eq:spectral_distance}
\end{equation}
A value of $d_{\mathrm{spec}} = 0$ indicates identical normalized spectra, meaning the manifolds have the same distribution of variation across scales. This measures spectral \emph{complexity} similarity without regard to eigenvector orientation.

\paragraph{Diagonal dominance.} For each spectral index $i$, the diagonal dominance is:
\begin{equation}
    \rho_i = \frac{C_{ii}^2}{\sum_{j=1}^{k_s} C_{ij}^2}.
    \label{eq:diag_dominance}
\end{equation}
If $\mathbf{C}$ is a permutation-free correspondence (i.e., the $i$-th mode in one manifold maps primarily to the $i$-th mode in the other), then $\rho_i \approx 1$. A mean $\bar{\rho} \ll 1$ indicates that spectral modes are scrambled across modalities: the $i$-th direction of variation in one representation space does not correspond to any single direction in the other. We report the mean $\bar{\rho} = \frac{1}{k_s}\sum_i \rho_i$.

\paragraph{Orthogonality deviation.} An isometric correspondence produces an orthogonal $\mathbf{C}$. We measure the deviation:
\begin{equation}
    \epsilon_{\mathrm{orth}} = \frac{1}{k_s}\|\mathbf{C}^{\top}\mathbf{C} - \mathbf{I}_{k_s}\|_F.
    \label{eq:ortho_error}
\end{equation}
A value of $\epsilon_{\mathrm{orth}} = 0$ indicates a perfectly isometric correspondence. Large values signal that the map is non-isometric---it stretches, compresses, or collapses spectral directions, meaning the two manifolds are not related by a distance-preserving transformation in spectral space.

\paragraph{Interpretation.} These three quantities decompose cross-modal compatibility into independent aspects. Two manifolds may have similar complexity ($d_{\mathrm{spec}} \approx 0$) but misaligned orientations ($\bar{\rho} \ll 1$), or aligned orientations but different complexities. The functional map framework requires \emph{all three} to be favorable: similar spectra, high diagonal dominance, and low orthogonality error. When one or more conditions fail, the diagnostics indicate which aspect of the representation geometry is incompatible, providing guidance for future methods.

\subsection{Baseline Methods}
\label{sec:baselines}

We compare against four methods, spanning the range from no alignment to full joint training.

\paragraph{Raw cosine similarity.} Truncates both feature matrices to $\min(d_v, d_t)$ dimensions and computes cosine similarity. Since the encoders are independently trained, their embedding dimensions carry no shared semantics; this baseline establishes the chance-level floor.

\paragraph{Orthogonal Procrustes~\cite{schonemann1966generalized}.} Given anchor pairs $(i, j) \in S$, computes $\mathbf{R}^* = \arg\min_{\mathbf{R}^{\top}\mathbf{R}=\mathbf{I}} \|\mathbf{Z}^v_S\mathbf{R} - \mathbf{Z}^t_S\|_F^2$ via the SVD of $(\mathbf{Z}^v_S)^{\top}\mathbf{Z}^t_S$. Features are truncated to $\min(d_v, d_t)$ dimensions before alignment.

\paragraph{Relative representations~\cite{moschella2023relative}.} Each point is re-represented by its cosine similarities to the $|S|$ anchor points within its own modality: $\mathbf{r}_i^m = [\cos(\mathbf{z}_i^m, \mathbf{z}_{s_1}^m), \ldots, \cos(\mathbf{z}_i^m, \mathbf{z}_{s_{|S|}}^m)]$. Cross-modal comparison is performed in this $|S|$-dimensional anchor-similarity space, which is modality-invariant by construction.

\paragraph{CLIP~\cite{radford2021learning}.} A jointly trained vision--language model, included as a strong supervised reference. CLIP ViT-B/32 was trained on 400 million image--text pairs with a contrastive objective. It represents the performance achievable with large-scale paired cross-modal supervision.

% =============================================================================
% Sections 4--5: Experiments and Discussion
% To be included in main.tex via \input{sections_4_5}
% =============================================================================

% =============================================================================
% 4. EXPERIMENTS
% =============================================================================

\section{Experiments}

\subsection{Setup}
\label{sec:setup}

\paragraph{Dataset.} We evaluate on Flickr30k~\cite{young2014image}, using 1{,}000 images from the test split, each annotated with five English captions (5{,}000 captions total). For methods that operate at the image level (all except CLIP), we represent each image's text by the mean of its five caption embeddings.

\paragraph{Encoders.} We use two independently pretrained encoders with no shared training signal:
\begin{itemize}
    \item \textbf{Vision:} DINOv2 ViT-B/14~\cite{oquab2024dinov2}, a self-supervised vision transformer producing 768-dimensional representations.
    \item \textbf{Text:} all-MiniLM-L6-v2~\cite{reimers2019sentence}, a distilled Sentence-BERT model producing 384-dimensional representations.
\end{itemize}
Neither model was exposed to paired image--text data during pretraining. We additionally test all-mpnet-base-v2 (768-dimensional) as a second text encoder in Experiments~4 and~5.

\paragraph{Hyperparameters.} For the spectral pipeline: $k$-NN graph with $k{=}15$, adaptive Gaussian bandwidth, spectral truncation $k_s{=}50$ (ablated in Experiment~2), ZoomOut refinement from $k_s{=}50$ to $k_{\mathrm{max}}{=}100$ in five steps. For the functional map optimization (Eq.~\ref{eq:fmap_opt}): $\lambda_1{=}0.1$, $\lambda_2{=}0.001$. Anchor pairs are selected uniformly at random; results are reported for a single random seed.

\paragraph{Metrics.} We report Recall@$K$ (R@$K$) for $K \in \{1, 5, 10\}$ in both directions: image-to-text (i2t) and text-to-image (t2i). For training-free methods operating at the image level, retrieval is evaluated over the $N{=}1{,}000$ image--text pairs; each image's five captions are treated as equivalent targets.\footnote{For non-CLIP methods, similarity is computed at the image level, so each image's five captions share the same score. Under this protocol, i2t caption-space R@$K$ reduces to image-space R@$\lceil K/5 \rceil$; therefore i2t R@1 and R@5 are identical because $\lceil 1/5 \rceil = \lceil 5/5 \rceil = 1$. We report both for comparability with standard benchmarks but focus discussion on R@1 and R@10.}

% ---- Experiment 1: Core Retrieval ----

\subsection{Experiment 1: Cross-Modal Retrieval}
\label{sec:exp1}

\begin{figure*}[t]
    \centering
    \includegraphics[width=\textwidth]{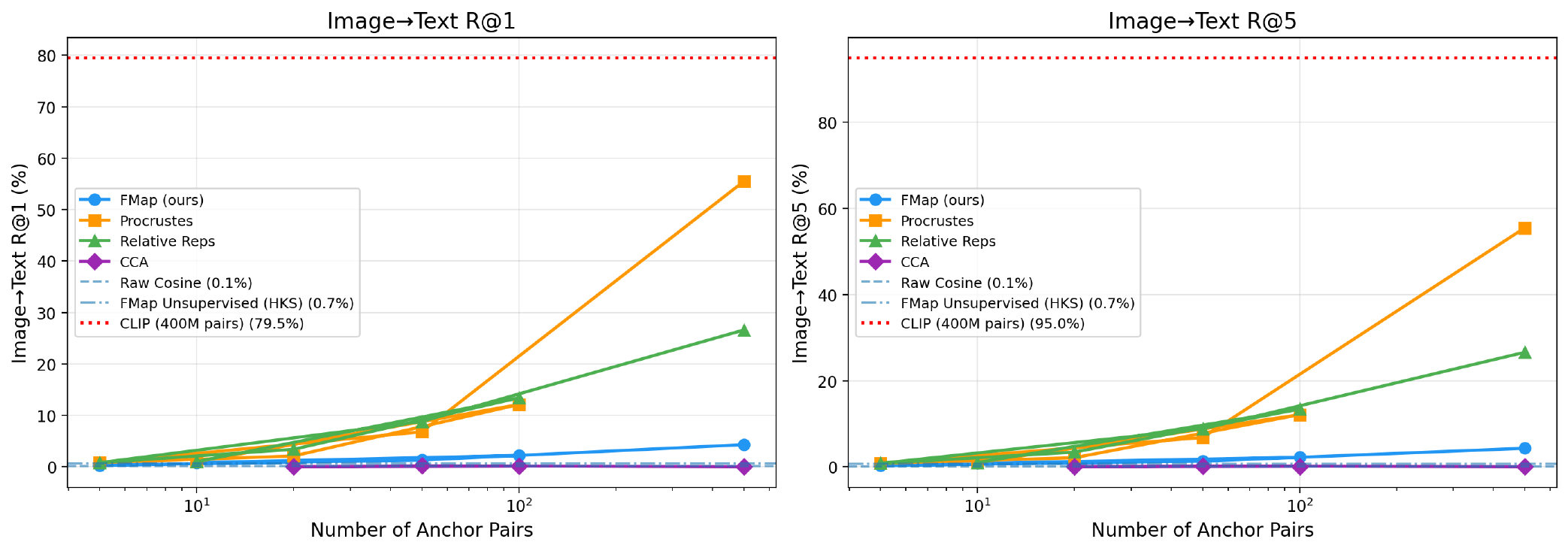}
    \caption{Image-to-text R@1 and R@5 as a function of anchor budget $|S|$ (log scale). The functional map (blue) improves with more anchors but grows substantially slower than Procrustes (orange) and relative representations (green). The CLIP reference line (red, dashed) indicates the performance achievable with full joint training on 400M pairs. CCA (purple) fails across all budgets.}
    \Description{Line plot of image-to-text recall versus anchor budget, comparing functional maps, Procrustes, relative representations, CCA, and a CLIP reference line.}
    \label{fig:anchor_budget}
\end{figure*}

Table~\ref{tab:main_results} presents representative operating points from the central comparison, alongside the zero-supervision baselines (raw cosine, unsupervised HKS) and the jointly trained CLIP model. The full six-budget sweep ($|S| \in \{5, 10, 20, 50, 100, 500\}$) is shown in Figure~\ref{fig:anchor_budget}.

\begin{table*}[t]
\centering
\caption{Image--text retrieval on Flickr30k (1{,}000 images, 5{,}000 captions). R@$K$ (\%) for image-to-text (i2t) and text-to-image (t2i). All training-free methods use the same DINOv2 + MiniLM encoder pair with $k_s{=}50$, ZoomOut refinement to $k_{\mathrm{max}}{=}100$, and a single random seed. CCA is omitted at $|S|{<}20$ (insufficient anchors). Bold indicates best training-free result per column; CLIP is shown as a strong supervised reference.}
\label{tab:main_results}
\small
\begin{tabular}{ll rrr rrr}
\toprule
& & \multicolumn{3}{c}{\textbf{Image $\to$ Text}} & \multicolumn{3}{c}{\textbf{Text $\to$ Image}} \\
\cmidrule(lr){3-5} \cmidrule(lr){6-8}
\textbf{Method} & $|S|$ & R@1 & R@5 & R@10 & R@1 & R@5 & R@10 \\
\midrule
Raw Cosine & 0 & 0.1 & 0.1 & 0.1 & 0.1 & 0.5 & 1.0 \\
FMap Unsupervised (HKS) & 0 & 0.7 & 0.7 & 1.1 & 0.4 & 1.7 & 3.2 \\
\midrule
FMap (ours) & 20 & 0.9 & 0.9 & 2.1 & 0.8 & 4.5 & 8.1 \\
Procrustes & 20 & 2.1 & 2.1 & 3.2 & 2.0 & 5.7 & 10.4 \\
Relative Reps & 20 & \textbf{3.4} & \textbf{3.4} & \textbf{5.0} & \textbf{3.5} & \textbf{9.3} & \textbf{14.1} \\
CCA & 20 & 0.0 & 0.0 & 0.3 & 0.1 & 0.4 & 1.0 \\
\midrule
FMap (ours) & 100 & 2.2 & 2.2 & 3.9 & 1.9 & 9.3 & 15.0 \\
Procrustes & 100 & 12.1 & 12.1 & \textbf{15.9} & 10.8 & 23.0 & 32.9 \\
Relative Reps & 100 & \textbf{13.4} & \textbf{13.4} & \textbf{19.0} & \textbf{12.1} & \textbf{27.5} & \textbf{36.7} \\
CCA & 100 & 0.1 & 0.1 & 0.2 & 0.1 & 0.5 & 0.9 \\
\midrule
FMap (ours) & 500 & 4.3 & 4.3 & 8.9 & 6.1 & 17.9 & 25.9 \\
Procrustes & 500 & \textbf{55.5} & \textbf{55.5} & \textbf{63.1} & \textbf{54.4} & \textbf{69.5} & \textbf{78.0} \\
Relative Reps & 500 & 26.6 & 26.6 & 34.6 & 26.7 & 50.9 & 62.3 \\
CCA & 500 & 0.0 & 0.0 & 0.1 & 0.1 & 0.4 & 0.9 \\
\midrule
\rowcolor[gray]{0.93}
CLIP ViT-B/32 & 400M & 79.5 & 95.0 & 98.1 & 58.8 & 83.4 & 90.0 \\
\bottomrule
\end{tabular}
\end{table*}

Three observations emerge from these results.

First, the functional map consistently underperforms Procrustes and relative representations across all anchor budgets. At $|S|{=}20$, the gap is moderate (FMap 0.9\% vs.\ Procrustes 2.1\% i2t R@1, a $2.3{\times}$ factor). At $|S|{=}500$, the gap becomes severe (FMap 4.3\% vs.\ Procrustes 55.5\%, a $12.9{\times}$ factor). The performance ratio \emph{worsens} with more supervision, indicating that additional anchor information benefits ambient-space methods far more than the spectral approach.

Second, the unsupervised functional map (HKS, zero anchors) achieves 0.7\% i2t R@1, which exceeds the raw cosine baseline (0.1\%) by a factor of seven. This confirms that the spectral bases do carry \emph{some} cross-modal information, but not enough for practical retrieval.

Third, CCA fails uniformly across all settings, with performance near the random baseline. This is consistent with the known sensitivity of CCA to the ratio of samples to dimensions: with $|S| \leq 500$ anchors and $d{=}384$ dimensions, the CCA solution is poorly conditioned.

% ---- Experiment 2: Spectral Dimension Ablation ----

\subsection{Experiment 2: Effect of Spectral Dimension}
\label{sec:exp2}

\begin{figure}[t]
    \centering
    \includegraphics[width=\columnwidth]{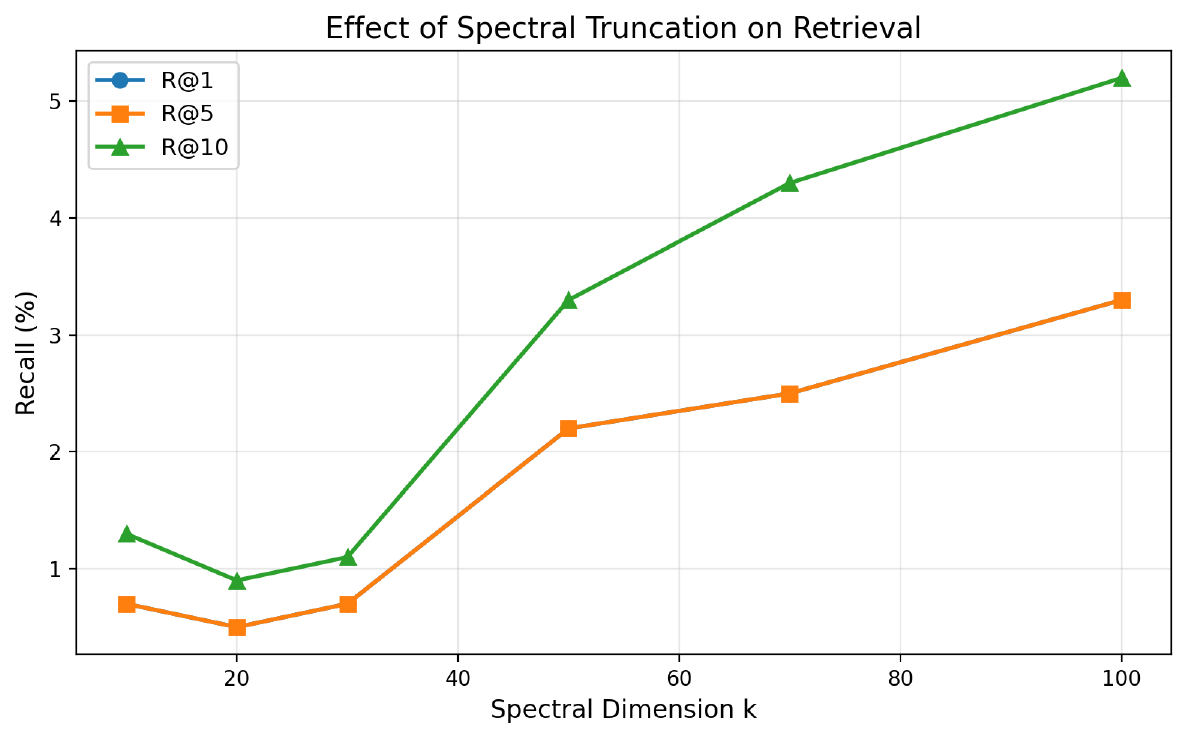}
    \caption{Image-to-text recall as a function of spectral dimension $k_s$. Performance increases with $k_s$ but remains an order of magnitude below ambient-space baselines at all truncations.}
    \Description{Line plot showing image-to-text recall as spectral dimension increases, with gradual gains that remain below ambient-space baselines.}
    \label{fig:k_ablation}
\end{figure}

Table~\ref{tab:k_ablation} and Figure~\ref{fig:k_ablation} show retrieval performance as a function of the spectral truncation $k_s$, with a fixed anchor budget of $|S|{=}50$. In this ablation, we disable ZoomOut to isolate the effect of $k_s$.

\begin{table}[t]
\centering
\caption{Effect of spectral dimension $k_s$ on functional map retrieval (R@$K$, \%, $|S|{=}50$ anchors) without ZoomOut refinement. Higher $k_s$ improves i2t retrieval monotonically but does not improve t2i, suggesting a floor imposed by eigenvector misalignment.}
\label{tab:k_ablation}
\small
\begin{tabular}{r rrr rrr}
\toprule
& \multicolumn{3}{c}{\textbf{i2t}} & \multicolumn{3}{c}{\textbf{t2i}} \\
\cmidrule(lr){2-4} \cmidrule(lr){5-7}
$k_s$ & R@1 & R@5 & R@10 & R@1 & R@5 & R@10 \\
\midrule
10  & 0.7 & 0.7 & 1.3 & 0.6 & 2.9 & 4.5 \\
20  & 0.5 & 0.5 & 0.9 & 0.1 & 0.8 & 1.2 \\
30  & 0.7 & 0.7 & 1.1 & 0.1 & 1.0 & 1.7 \\
50  & 2.2 & 2.2 & 3.3 & 0.1 & 0.4 & 1.0 \\
70  & 2.5 & 2.5 & 4.3 & 0.0 & 0.5 & 0.9 \\
100 & 3.3 & 3.3 & 5.2 & 0.1 & 0.5 & 0.9 \\
\bottomrule
\end{tabular}
\end{table}

Image-to-text R@1 increases monotonically from 0.7\% ($k_s{=}10$) to 3.3\% ($k_s{=}100$), confirming that higher spectral resolution captures more cross-modal signal. However, even at $k_s{=}100$---the maximum computed in this ablation---the performance remains far below Procrustes at the same anchor budget (Figure~\ref{fig:anchor_budget}). The bottleneck is not the number of spectral modes retained but the quality of the correspondence between them.

A notable asymmetry appears in the text-to-image direction: t2i performance does not improve with $k_s$ and in fact slightly decreases for $k_s \geq 20$. We hypothesize that higher-frequency spectral components introduce noise from modality-specific structure that harms the text-to-image direction more than it helps.

% ---- Experiment 3: Spectral Diagnostics ----

\subsection{Experiment 3: Spectral Diagnostics}
\label{sec:exp3}

\begin{figure*}[t]
    \centering
    \includegraphics[width=\textwidth]{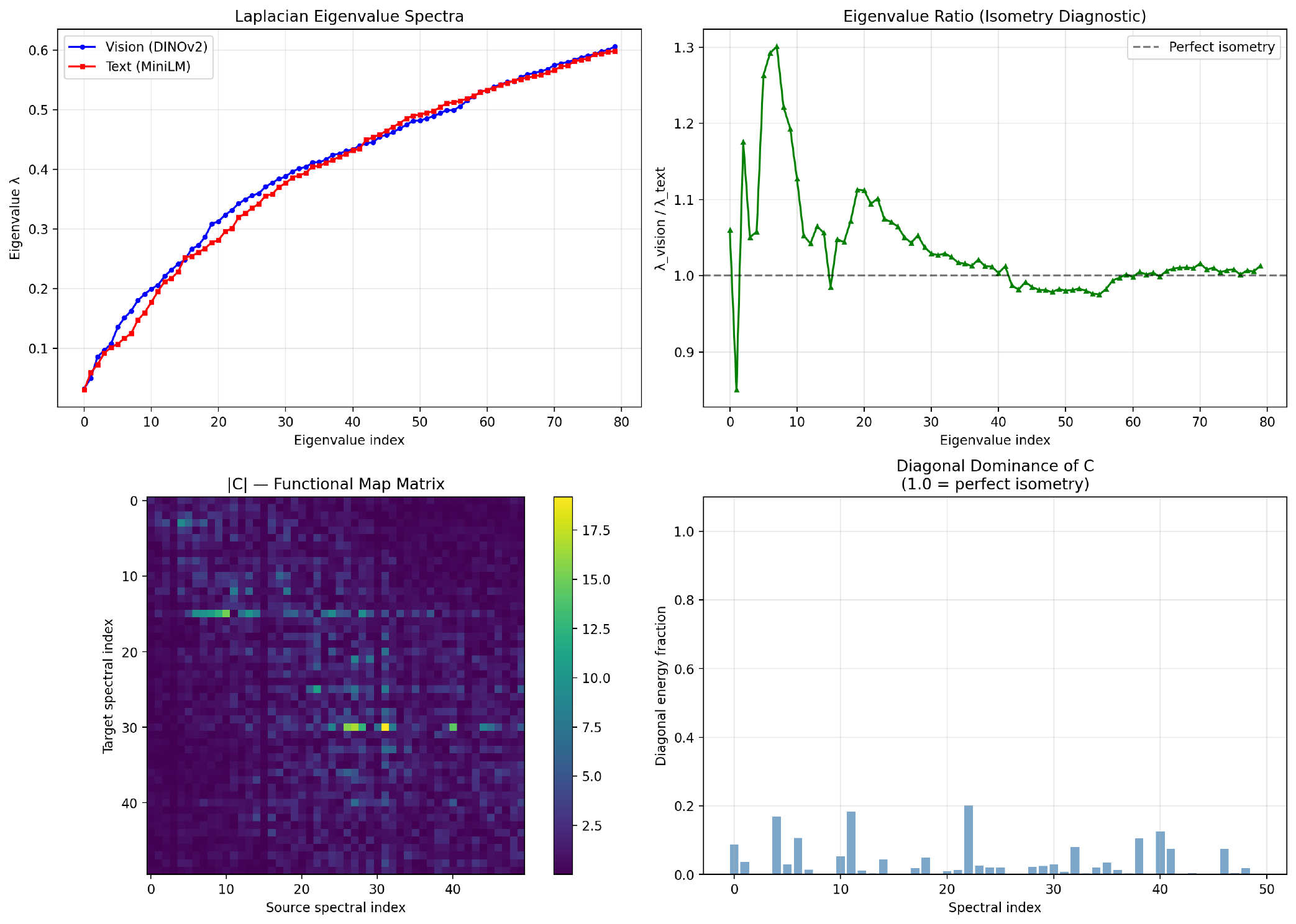}
    \caption{Spectral diagnostics for the DINOv2--MiniLM encoder pair. \textbf{Top left:} Laplacian eigenvalue spectra are quantitatively similar. \textbf{Top right:} eigenvalue ratio $\lambda_i^v / \lambda_i^t$ deviates from 1.0 primarily at low frequencies (indices 0--10), then stabilizes. \textbf{Bottom left:} absolute values of the functional map matrix $|\mathbf{C}|$; energy is scattered across off-diagonal bands (rows $\sim$15 and $\sim$30) rather than concentrated on the diagonal. \textbf{Bottom right:} diagonal energy fraction $\rho_i$ per spectral index; all values are below 0.2, most below 0.05.}
    \Description{Four-panel diagnostic figure showing similar Laplacian spectra, frequency-wise eigenvalue ratios, an off-diagonal functional map matrix magnitude heatmap, and low diagonal energy fractions across spectral indices.}
    \label{fig:spectral_diag}
\end{figure*}

This experiment examines the internal structure of the spectral bases and the functional map, using the diagnostic quantities defined in \S\ref{sec:diagnostics}. Figure~\ref{fig:spectral_diag} presents the four diagnostic panels. Table~\ref{tab:diagnostics} summarizes the aggregate quantities.

\begin{table}[t]
\centering
\caption{Spectral diagnostic quantities for the DINOv2 (vision) and MiniLM (text) encoder pair, computed on $N{=}1{,}000$ Flickr30k samples with $k_s{=}50$ and $|S|{=}50$ anchors. For reference, values typical of near-isometric shape correspondence are shown in the right column~\cite{ovsjanikov2012functional, melzi2019zoomout}.}
\label{tab:diagnostics}
\small
\begin{tabular}{l r r}
\toprule
\textbf{Diagnostic} & \textbf{Observed} & \textbf{Shape matching} \\
\midrule
Spectral distance $d_{\mathrm{spec}}$ & 0.043 & $< 0.01$ \\
Mean diagonal dominance $\bar{\rho}$ & $< 0.05$ & $> 0.7$ \\
Orthogonality error $\epsilon_{\mathrm{orth}}$ & 70.15 & $< 0.1$ \\
\midrule
Eigenvalue range (vision) & $[0.032, 0.662]$ & --- \\
Eigenvalue range (text) & $[0.030, 0.655]$ & --- \\
\bottomrule
\end{tabular}
\end{table}

The eigenvalue spectra (Figure~\ref{fig:spectral_diag}, top left) are strikingly similar: both follow the same concave growth profile from ${\sim}0.03$ to ${\sim}0.66$, with a normalized spectral distance of just 0.043. The eigenvalue ratio (top right) deviates from 1.0 primarily at the lowest frequencies---indices 0--10 show ratios up to 1.3---and converges toward 1.0 at higher frequencies. This indicates that the coarsest semantic structure (captured by the lowest eigenvectors) shows the most inter-modal variation, while finer-grained structure is more spectrally compatible.

The functional map matrix (bottom left) reveals the critical failure. In successful shape correspondence, $|\mathbf{C}|$ is approximately diagonal, with the $i$-th row dominated by the $(i,i)$ entry. Here, energy is concentrated in horizontal bands around rows 15 and 30, indicating that multiple source spectral modes map to the same small set of target modes. The diagonal dominance plot (bottom right) confirms this quantitatively: no spectral index achieves $\rho_i > 0.2$, and the mean $\bar{\rho}$ is below 0.05.

The orthogonality error $\epsilon_{\mathrm{orth}} = 70.15$ confirms that $\mathbf{C}$ is not close to orthogonal. For comparison, functional maps between near-isometric shapes typically yield $\epsilon_{\mathrm{orth}} < 0.1$~\cite{ovsjanikov2012functional}. The value observed here is three orders of magnitude larger, indicating that the correspondence between the two representation manifolds is far from isometric.

% ---- Experiment 4: Composability ----

\subsection{Experiment 4: Composability}
\label{sec:exp4}

Table~\ref{tab:composability} evaluates the composability property of functional maps. We compute separate maps from DINOv2 to MiniLM ($\mathbf{C}^{v \to t_1}$) and from MiniLM to mpnet ($\mathbf{C}^{t_1 \to t_2}$), each using 20 anchor pairs drawn independently. The composed map $\mathbf{C}^{v \to t_2}_{\mathrm{comp}} = \mathbf{C}^{t_1 \to t_2} \cdot \mathbf{C}^{v \to t_1}$ is compared against a direct map $\mathbf{C}^{v \to t_2}_{\mathrm{direct}}$ computed from 20 anchor pairs of DINOv2--mpnet.

\begin{table}[t]
\centering
\caption{Composability evaluation. The composed map (DINOv2$\to$MiniLM$\to$mpnet) uses no direct DINOv2--mpnet anchor pairs, while the direct map uses 20.}
\label{tab:composability}
\small
\begin{tabular}{l rrr}
\toprule
& \multicolumn{3}{c}{\textbf{i2t R@$K$ (\%)}} \\
\cmidrule(lr){2-4}
\textbf{Method} & R@1 & R@5 & R@10 \\
\midrule
Composed ($v{\to}t_1{\to}t_2$) & 0.3 & 0.3 & 0.7 \\
Direct ($v{\to}t_2$, 20 anchors) & 1.3 & 1.3 & 2.1 \\
\midrule
Random baseline & 0.1 & 0.5 & 1.0 \\
\bottomrule
\end{tabular}
\end{table}

The composed map achieves 0.3\% i2t R@1, compared to 1.3\% for the direct map. Both exceed the random baseline (0.1\%), confirming that the composition mechanism transmits \emph{some} cross-modal information. However, the composed map is $4.3{\times}$ worse than the direct map. Since composition error is multiplicative---the error in $\mathbf{C}^{v \to t_2}_{\mathrm{comp}}$ is bounded by the product of the individual map errors~\cite{ovsjanikov2012functional}---this degradation is expected when both individual maps are already poor. The composability mechanism is mathematically sound; it is the individual map quality that limits the composed result.

% ---- Experiment 5: Encoder Pair Variation ----

\subsection{Experiment 5: Encoder Pair Variation}
\label{sec:exp5}

To verify that our findings are not specific to the MiniLM text encoder, Table~\ref{tab:encoder_pairs} reports functional map retrieval for two encoder pairings using $|S|{=}50$ anchors and ZoomOut refinement.

\begin{table}[t]
\centering
\caption{Functional map retrieval across encoder pairings ($|S|{=}50$ anchors, $k_s{=}50$, ZoomOut refinement to $k_{\mathrm{max}}{=}100$).}
\label{tab:encoder_pairs}
\small
\begin{tabular}{ll rrr}
\toprule
\textbf{Vision} & \textbf{Text} & \multicolumn{3}{c}{\textbf{i2t R@$K$ (\%)}} \\
\cmidrule(lr){3-5}
& & R@1 & R@5 & R@10 \\
\midrule
DINOv2-B & MiniLM & 1.7 & 1.7 & 3.3 \\
DINOv2-B & mpnet & 1.0 & 1.0 & 2.3 \\
\bottomrule
\end{tabular}
\end{table}

Both pairings yield comparable results in the low single digits, confirming that the performance limitation is not an artifact of a particular encoder choice. The mpnet encoder (768-dimensional, same as DINOv2) produces marginally lower performance than MiniLM (384-dimensional), suggesting that matching dimensionality does not help---the mismatch is geometric, not dimensional.

% =============================================================================
% 5. DISCUSSION
% =============================================================================

\section{Discussion}

\subsection{The Spectral Complexity--Orientation Gap}

The central finding of this work is the decoupling of two properties that are linked in shape correspondence but independent in cross-modal neural representations.

In shape matching, near-isometric shapes share both eigenvalue spectra and eigenvector correspondence. This linkage is a theorem: if two Riemannian manifolds are related by an isometry, their Laplace--Beltrami operators are unitarily equivalent, which implies identical eigenvalues and related eigenfunctions~\cite{ovsjanikov2012functional}. The entire functional map framework depends on this linkage.

Our experiments reveal that for independently pretrained neural encoders, the eigenvalue half of this linkage holds approximately (spectral distance $= 0.043$) but the eigenvector half does not (diagonal dominance $< 0.05$, orthogonality error $= 70.15$). We term this the \emph{spectral complexity--orientation gap}. It means:

\begin{itemize}
    \item The two representation manifolds have similar intrinsic complexity---they capture a comparable number of directions of variation at each scale. This is consistent with the Platonic Representation Hypothesis~\cite{huh2024platonic}: both models, trained on different data modalities, converge to representations that parse the world into a similar number of independent factors.

    \item The axes along which this variation is organized are completely different. The first eigenvector of the vision manifold (the coarsest mode of visual variation) does not correspond to any single mode of textual variation. Instead, it maps to a diffuse mixture of many textual modes.
\end{itemize}

This gap is not a limitation of the functional map computation. It is a structural property of the representations themselves. Increasing the anchor budget, changing the spectral truncation, or switching text encoders does not close it (Tables~\ref{tab:main_results},~\ref{tab:k_ablation}, and~\ref{tab:encoder_pairs}). The Laplacian commutativity regularization in Eq.~\ref{eq:fmap_opt}---which biases $\mathbf{C}$ toward frequency-preserving maps---actively harms retrieval in this setting because the assumption it encodes (that low-frequency visual structure corresponds to low-frequency textual structure) is empirically false.

\subsection{Why Ambient-Space Methods Outperform Spectral Methods}

Procrustes alignment at $|S|{=}500$ achieves 55.5\% i2t R@1---a factor of $12.9{\times}$ over the functional map. This gap has a precise explanation.

Procrustes operates in the full $d$-dimensional embedding space and finds the global rotation minimizing anchor reconstruction error. It makes no assumption about intrinsic manifold geometry, treating alignment as an extrinsic point-cloud problem. Because embeddings are high-dimensional ($d{=}384$), Procrustes has many degrees of freedom ($d \times d$ parameters, constrained to $d(d{-}1)/2$ by orthogonality) to fit anchor correspondences.

The functional map, by contrast, projects to a $k_s$-dimensional spectral basis ($k_s{=}50$ to $100$) and solves for a $k_s \times k_s$ map in that compressed space. This projection discards information present in ambient features. In shape matching, discarded content is often high-frequency noise and low-frequency components retain semantics. In cross-modal neural representations, the opposite appears true: useful cross-modal signal is not concentrated in low frequencies, so projection becomes a lossy bottleneck rather than a helpful filter.

Relative representations outperform Procrustes at lower anchor budgets ($|S| \leq 100$) because they build a modality-invariant, non-parametric coordinate system via anchor relations rather than fitting a transformation. This is more data-efficient but saturates earlier: at $|S|{=}500$, Procrustes (55.5\% R@1) overtakes relative representations (26.6\% R@1), likely because the rotation becomes well-conditioned with enough anchors.

\subsection{Eigenvalue Convergence as Evidence for the Platonic Representation Hypothesis}

While the negative retrieval result dominates the practical conclusions, the eigenvalue convergence finding (Table~\ref{tab:diagnostics}, Figure~\ref{fig:spectral_diag}) has independent scientific value.

Prior evidence for the Platonic Representation Hypothesis~\cite{huh2024platonic} has relied on CKA~\cite{kornblith2019similarity} and kernel alignment, which measure global similarity between representation geometries without decomposing that similarity by scale. Our spectral analysis provides a complementary perspective: the eigenvalue spectrum of a graph Laplacian captures how the representation manifold distributes its variation across scales. The finding that DINOv2 and MiniLM have nearly identical normalized spectra (distance $= 0.043$) means they not only represent similar total structure, but allocate it similarly across coarse-to-fine levels.

This is a stronger statement than high CKA alone. Two representations could have high CKA with different spectral profiles if their global kernel structures happen to align despite different scale distributions. Conversely, the eigenvalue convergence we observe implies a specific structural similarity: the ``bandwidth'' of the representation---how many independent directions of variation it supports at each granularity---is consistent across modalities.

We note two caveats. First, the spectral distance is computed on a finite sample ($N{=}1{,}000$) and is subject to estimation error in the graph Laplacian. Second, the similarity may partly reflect shared properties of the $k$-NN graph construction rather than deep properties of the representations. Evaluating with larger $N$ and alternative graph constructions would help disentangle these factors.

\subsection{Limitations}

We identify four limitations of the present study.

\textit{Scale.} We evaluate on 1{,}000 images. This is enough for stable spectral diagnostics, but retrieval metrics remain noisy; full Flickr30k-scale evaluation would give more reliable comparisons.

\textit{Encoder diversity.} We test one vision encoder (DINOv2-B) and two text encoders (MiniLM, mpnet). The spectral complexity--orientation gap may differ for larger models, different training objectives, or partially aligned vision--language models.

\textit{Graph construction sensitivity.} The spectral basis depends on the $k$-NN graph and kernel bandwidth, but we use one setting ($k{=}15$, adaptive bandwidth). Broader graph-construction sweeps could change the diagnostics.

\textit{Scope of the negative result.} We test functional maps on \emph{independently} pretrained encoders. The method may work better when encoders already share structure (e.g., overlapping pretraining or light alignment), so our result is about limits of purely post-hoc spectral alignment, not functional maps in general.

% =============================================================================
% Section 6: Conclusion
% =============================================================================

\section{Conclusion}

We applied the functional map framework from computational geometry to training-free cross-modal alignment between independently pretrained vision and language encoders. The framework underperforms ambient-space baselines for retrieval---Procrustes and relative representations achieve $5{\times}$ to $13{\times}$ higher Recall@1 across all anchor budgets tested---but its diagnostic value is the principal contribution. The spectral analysis exposes a structural property we term the \emph{spectral complexity--orientation gap}: the graph Laplacian eigenvalue spectra of DINOv2 and MiniLM are quantitatively similar (normalized distance $= 0.043$), yet their eigenvector bases are effectively unaligned (diagonal dominance $< 0.05$, orthogonality error $= 70.15$). This decoupling marks a precise boundary condition for spectral methods in multimodal alignment and offers a finer-grained characterization of cross-modal representation geometry than global measures such as CKA~\cite{kornblith2019similarity}.

The gap points to two directions for future work. First, \emph{spectral alignment}---finding a rotation in spectral space that brings eigenvector bases into correspondence without modifying the underlying representations---would make the functional map framework applicable; whether such a rotation exists and can be computed efficiently is an open problem, conceptually analogous to unsupervised cross-lingual alignment~\cite{conneau2018word} but in the spectral domain. Second, the diagnostic quantities we introduce (spectral distance, diagonal dominance, orthogonality error) could serve as \emph{model selection criteria}: computing them before attempting alignment may predict which method is appropriate for a given encoder pair, a hypothesis that requires evaluation across a wider range of architectures and training procedures than we examine here.
% =============================================================================
% BIBLIOGRAPHY
% =============================================================================

\bibliographystyle{ACM-Reference-Format}
\bibliography{references}

\end{document}